\documentclass{article} 
\usepackage{iclr2024_conference,times}

\usepackage{amsmath,amsfonts,bm}

\def\eqref#1{equation~\ref{#1}}

\def\1{\bm{1}}

\DeclareMathAlphabet{\mathsfit}{\encodingdefault}{\sfdefault}{m}{sl}
\SetMathAlphabet{\mathsfit}{bold}{\encodingdefault}{\sfdefault}{bx}{n}

\usepackage{hyperref}
\usepackage{url}

\usepackage[utf8]{inputenc}
\usepackage{xspace}
\usepackage{graphicx}
\usepackage{booktabs}
\usepackage{multirow}

\usepackage{footnote}
\makesavenoteenv{tabular}
\makesavenoteenv{table}

\newcommand{\aitw}{\textsc{AitW}\xspace}
\newcommand{\datasetgoogle}{\textsc{GoogleApps}\xspace}
\newcommand{\datasetinstall}{\textsc{Install}\xspace}
\newcommand{\datasetshopping}{\textsc{WebShopping}\xspace}
\newcommand{\datasetgeneral}{\textsc{General}\xspace}
\newcommand{\datasetsingle}{\textsc{Single}\xspace}

\iclrfinalcopy 
\title{Training a Vision Language Model as \\ Smartphone Assistant}

\author{
Nicolai Dorka \textsuperscript{1,2} ~~~~~~~ Janusz Marecki \textsuperscript{1} ~~~~~~~ Ammar Anwar \textsuperscript{1}\\
~\textsuperscript{1}Agile Loop ~~~~~ \textsuperscript{2}University of Freiburg \\
~Correspondence to~\texttt{nicolai.dorka1@gmail.com}
}

\begin{document}

\maketitle

\begin{abstract}
Addressing the challenge of a digital assistant capable of executing a wide array of user tasks, our research focuses on the realm of instruction-based mobile device control.
We leverage recent advancements in large language models (LLMs) and present a visual language model (VLM) that can fulfill diverse tasks on mobile devices.
Our model functions by interacting solely with the user interface (UI). It uses the visual input from the device screen and mimics human-like interactions, encompassing gestures such as tapping and swiping.
This generality in the input and output space allows our agent to interact with any application on the device.
Unlike previous methods, our model operates not only on a single screen image but on vision-language sentences created from sequences of past screenshots along with corresponding actions.
Evaluating our method on the challenging Android in the Wild benchmark
demonstrates its promising efficacy and potential.
\end{abstract}

\section{Introduction}

As mobile devices continue to evolve, there is an increasing demand for intuitive and efficient methods of interaction.
Traditionally, users operate their devices through a series of taps and gestures on the screen. 
However, in many instances it is more natural and convenient to express commands in natural language rather than directly interacting with the device. 
A digital assistant fulfilling such tasks holds particular promise for demographics such as older individuals unfamiliar with technology, those with physical limitations, and for situations where direct interaction with the device is not feasible.
Thus, the development of mobile device-control systems capable of comprehending and executing natural language instructions holds significant value.

The great success of large language models (LLMs) 
\citep{achiam2023gpt,team2023gemini}
in recent years has nourished approaches using LLMs to create helpful digital assistants able to fulfill user-specified instructions \citep{autogpt,agentgpt}.
While the capabilities of such methods are remarkable their utility remains severely constrained in various scenarios.
Many existing methods rely on LLMs to issue API calls to applications. However, this strategy has several limitations. Notably, not all applications offer APIs, and even when available, integrating multiple API languages poses challenges either in terms of training or the context length of the LLM.
Furthermore, summarizing transitions to new states in language, particularly when incorporating visual or spatial information, presents further hurdles.
A more natural approach lies in controlling devices through the user interface (UI), mirroring human interaction patterns. The UI provides comprehensive information in a standardized format across applications, simplifying the control process. Actions within the UI, such as tapping or swiping, remain consistent, facilitating generalization across different applications.
The innate ability of native mobile users to effortlessly navigate unfamiliar applications underscores the feasibility of generalizing UI interactions across diverse applications.

Visual language models (VLMs) \citep{team2023gemini,gpt4vision} represent a promising avenue for device control exclusively through screen inputs.
Harnessing both visual and textual data, VLMs possess the dual capabilities of understanding user instructions in natural language and interpreting the current screen environment. Leveraging the generalization powers of LLMs, VLMs excel in formulating complex actions in text format.


In contrast to previous approaches \citep{zhan2023onlylookatscreens,hong2023cogagent}, we utilize a history of screenshots as input for our model. This is crucial in certain situations to help the model determine its next steps. Additionally, we create a method to translate the actions possible on mobile devices into a language format that the VLM can easily understand and generate.
We train the model on vision-language sentences created from the history of screen images and action transcriptions.
This allows us to train the VLM using techniques similar to those used for generating language, such as masked self-attention and next-token prediction.

To fine-tune the VLMs for controlling devices, we utilize the \aitw dataset \citep{rawles2023androidinthewild}, which contains expert demonstrations covering a wide range of mobile control tasks.
Our contributions additionally include the training of different base VLMs to gain insights into what is crucial for further refining them as agents for device control.
We experiment with a VLM that has been pretrained on vision-language tasks, as well as a new VLM developed by us that has not undergone such pretraining.
Our experiments indicate that pretraining on vision-language tasks is beneficial, particularly when OCR tasks are included.

Our best-performing model is a novel VLM capable of controlling mobile devices based on language commands using solely the UI.
We assess the effectiveness of our model by evaluating it on the \aitw benchmark.
Our analysis reveals state-of-the-art results, confirming the efficacy of our approach.

\section{Related Work}

\subsection{Large Language Models for Decision Making}
The surprisingly strong capabilities of LLMs have sparked interest in using them not just as writing assistants but also as agents for tasks specified by the user. 
LLMs were shown to be able to make use of existing tools \citep{parisi2022talm,schick2023toolformer,qin2023toolllm,patil2023gorilla}.
By formulating a feedback loop completely in natural language a powerful LLM can be used to complete various tasks by clever prompting. 
Famous examples are
AutoGPT \citep{autogpt}, AgentGPT \citep{agentgpt}, and BabyAGI \citep{babyagi}.
In robotics, LLMs have been successfully applied for instruction-based high-level task planning \citep{saycan}.
LLMs can also generate low-level robotic skills from instructions \citep{brohan2023rt2}.
Further, LLM agents have been successfully applied in Minecraft
\citep{lifshitz2023steve,wang2023voyager}.

\subsection{Instruction-based Mobile Device Control}

Several prior works studied the control of mobile devices given user instructions.
One recent line of work uses LLMs to do so by prompting the LLM in the right way and without actual training of the model.
The state information is represented in natural language by an additional vision module \citep{song2023navigating}, by summarizing the GUI in a simplified HTML representation \citep{wen2023empowering}, or by using page information from the Document Object Model (DOM) \citep{ding2024mobileagent}.
All of these approaches differ from our method in that
they involve a module that summarizes the state in natural language. Then, a general LLM, which is not specifically fine-tuned for the task, selects an action based on this summary.

Other very recent works use visual language models to control mobile devices. 
\citet{yan2023gpt4wonderland} use the powerful GPT4-V \citep{gpt4vision} and show that it has strong zero-shot GUI navigation capabilities with the right prompt.
While this shows the potential of VLMs for device control the performance is still limited as the model - different to ours - is not finetuned for this task.
Concurrently to our work, in SeeClick \citep{cheng2024seeclick} a VLM is trained with the specific purpose of GUI grounding which describes the ability to accurately locate screen elements. They collect a large dataset and train the model to predict the position of screen elements. Afterwards, the model can further be trained on device control data.
Another recent work proposed Auto-UI \citep{zhan2023onlylookatscreens} which is the first VLM trained for device control solely from screen inputs.
In contrast to our work, they use the BLIP-2 \citep{li2023blip2} vision encoder and fused self-attention. As a consequence, their method does not take a history of screenshots as input while our method can do so.
In the same way, CogAgent  \citep{hong2023cogagent} fine-tunes the 17-billion parameter CogVLM \citep{wang2023cogvlm} to control devices using screen images as input. Both CogAgent and Auto-UI differ from our approach in that they do not utilize a sequence of past screenshots as inputs.

\section{Method}

\begin{figure*}[t]
\centering 
\includegraphics[width=0.95\textwidth]{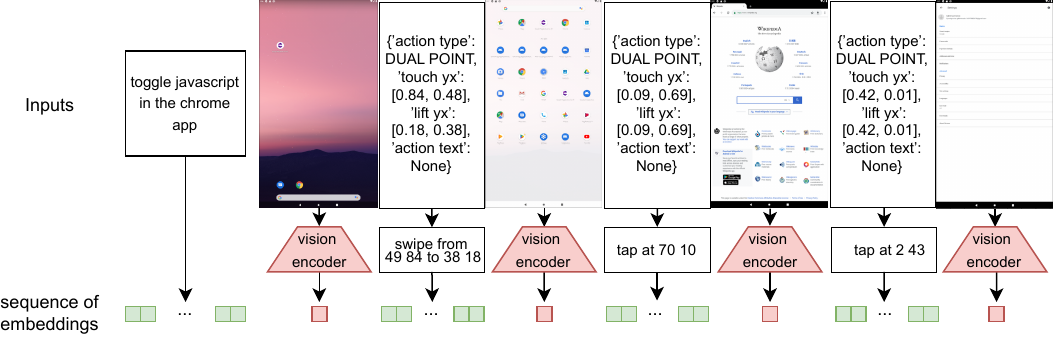}
\caption{Visualization of our approach. We create a sequence of embedding vectors for the VLM from the instruction, the history of screenshots, and the history of actions which are first translated to natural language which is then encoded into token embeddings. Depending on the vision encoder the number of vision embedding vectors can vary.}
\label{fig:approach}
\end{figure*}


Our goal is to develop an algorithm that can execute user-provided instructions expressed in natural language on mobile devices.
For instance, an instruction might be "clear the browsing history in Chrome". Notably, we operate under the assumption that access is restricted solely to the user interface (UI). This assumption is grounded in the reality that not all applications offer underlying APIs for direct control. Consequently, our algorithm is designed to navigate and manipulate the device exclusively through screen inputs, mirroring human interaction patterns. For computers, this involves simulating mouse clicks and keyboard inputs, while for mobile devices, it entails replicating touch gestures.
In the subsequent sections, we delineate the architecture of our device control algorithm.

\subsection{Vision Language Models}

We experiment with two kinds of VLMs.
The first one connects a pretrained language model with a pretrained vision encoder without training the resulting VLM on vision-language tasks before finetuning it as an agent.
The second approach uses an open-source VLM that has been pretrained on other vision-language tasks before.

\subsubsection{LLama + ViT}

We introduce a vision language model (VLM) \citep{team2023gemini,gpt4vision,driess2023palm}
tailored for device control within the UI environment. The VLM's primary function is to predict the subsequent action required to fulfill a given instruction. Its input comprises the instruction itself along with a historical record consisting of screenshots and associated actions with the latter being formatted in natural language. The mapping between these natural language representations and the corresponding device actions is detailed in Section \ref{sec:action_representation}.

To accommodate both textual and visual inputs, we encode both modalities into a unified embedding space. Token embeddings from the language model encode the textual components, while a vision encoder transforms each image into a lower-dimensional representation. A trainable linear projection aligns these visual embeddings with the dimensionality of the token embeddings. Subsequently, the language model operates on a sequence of these embeddings. The sequence begins with the token embeddings of the instruction, followed by the embeddings of the initial screen state, the first action, the subsequent screen state, and so forth. This approach enables us to represent a complete trajectory as a sequence of embeddings, facilitating seamless integration of text and visual information.
A visualization of this approach is depicted in Figure \ref{fig:approach}.

For the vision encoder, we leverage a pre-trained Vision Transformer (ViT) model \citep{dosovitskiy2020image}, configured with 320 million parameters, designed to process images of dimensions $ 384 \times 384$. The choice of the input resolution is deliberate, as it ensures the recognition of fine details such as small text present on the screen. The output of the vision encoder is projected into the token embedding space of the language model using a learnable projection matrix
\begin{equation}
    x_{embed} = W_{proj}~\text{VE}(x_{img}),
\end{equation}
where $x_{img}$ is the input image, $\text{VE}$ is the vision encoder mapping the image to an $l$-dimensional vector, and $W_{proj} \in \mathbb{R}^{k \times l}$ is a learnable projection matrix mapping the encoded image into the $l$-dimensional space of the token embeddings of the language model.

For the language model, we utilize LLama-2-7B \citep{touvron2023llama2}.
We opt for a decoder-only model to prioritize the generation of text auto-regressively, aligning with the requirement to predict actions in textual form. LLama-v2 models are renowned for their powers in text generation tasks, making them a natural fit for our application. The decision to utilize the 7B parameter version is motivated by computational efficiency considerations, although larger versions could potentially yield even more robust performance. Nonetheless, our experiments with the 7B parameter version demonstrate promising results, validating the effectiveness of our approach.

It is worth noting that, unlike previous approaches, our method incorporates complete historical states, enabling a more comprehensive understanding of the context and enhancing the robustness of our control mechanism.

\subsubsection{Qwen-VL}

The visual encoder of Qwen-VL \citep{bai2023qwenvl} utilizes the Vision Transformer (ViT) architecture with pre-trained weights from Openclip’s ViT-bigG \citep{openclip}. Input images are resized to a resolution of $ 448 \times 448$ and processed by the visual encoder, which splits them into patches with a stride of 14, generating image features.
To reduce the length of the feature sequence, Qwen-VL introduces a position-aware vision-language adapter. This adapter, initialized with a randomly initialized single-layer cross-attention module, compresses image features using trainable query vectors and positional encodings. The compressed feature sequence, fixed at length $256$, is then fed into the language model which is initialized with pre-trained weights from Qwen-7B \citep{bai2023qwen}.
In total, the model has $9.6$ billion parameters.

Qwen-VL has been pretrained on a diverse set of vision-language tasks.
In particular, this encompasses tasks such as optical character recognition (OCR) and tasks that involve locating objects within an image.
We suspect both to be helpful pretraining tasks for understanding screens as this requires understanding of the text as well as locating buttons.

\subsection{Action Representation for Mobile Devices}
\label{sec:action_representation}

We leverage the action space provided by the \aitw benchmark, which comprises four fields:
\emph{type}, \emph{touch\_point}, \emph{lift\_point} (exclusive for gesture actions), and \emph{typed\_text} (exclusive for typing actions). Within this framework, six distinct action types are defined: \emph{dual-point gesture}, \emph{type}, \emph{go\_back}, \emph{go\_home}, \emph{enter}, \emph{task\_complete}, and \emph{task\_impossible}.
The dual-point gesture is accompanied by the touch and lift parameters. 
It can either mean a swipe if touch and lift are distinct from each other or a tap if they are sufficiently similar. Both the touch and lift parameters are given by the $(x,y)$ coordinate on the screen where it is touched or lifted from.

We translate these actions into natural language by initially stating the action type, with additional specifications for \emph{dual-point gestures} to differentiate between taps and swipes. For tap actions, we include the coordinates of the touch point separated by a whitespace character, with coordinates discretized into bins ranging from $0$ to $99$. For example, a tap action might be represented as "tap at 7 90". Swipe actions are expressed as "swipe from 3 44 to 40 48".
Typing actions are depicted as "Input text \emph{"}\emph{txt}\emph{"}",
with \emph{txt} representing the \emph{typed\_text} for the typing action. The actions \emph{go\_back}, \emph{go\_home}, \emph{enter} are preceded by "press" to indicate a button press action, for instance, "press home".
The detailed mappings are shown in Table \ref{table:action_map} in the appendix.

\subsection{Training}
During the training of LLama+ViT, we freeze the vision encoder and token embeddings of the LLM, focusing training efforts solely on the language model itself and the projection matrix $W_{proj}$ used to map vision encoder outputs to the language model's embedding size.
For the Qwen-VL model, we freeze all parts of the network except the language model because the vision and vision projection parts are already pretrained to produce useful inputs to the language model.
To reduce computational and hardware demands, we employ LoRA \citep{hu2021lora} for both models.

Our training dataset comprises demonstrations, with the complete model trained via next-token prediction and masked self-attention. This approach mirrors traditional LLM fine-tuning on domain-specific datasets, with the exception that during loss calculation, predictions corresponding to the instruction and image embeddings are disregarded. Consequently, the LLM is trained to accurately predict tokens corresponding to the correct actions based on the provided instruction and the history of actions and screenshots so far.
In this way, the model is trained on complete trajectories including the instruction in the same way as if it would be trained on complete sentences.

\section{Experiments}

\subsection{Android in the Wild Dataset}
We utilize the Android in the Wild (\aitw)\citep{rawles2023androidinthewild}  benchmark dataset in our assessment. \aitw is a comprehensive benchmark dataset designed for UI control consisting of expert demonstrations encompassing natural language instructions, screenshots, and actions. 
The benchmark dataset comprises five subsets: \datasetgoogle, \datasetinstall, \datasetshopping, \datasetgeneral, and \datasetsingle, each serving different purposes.  Each subset is divided into training, validation, and test sets based on episodes, with proportions of 80\%/10\%/10\%.
We evaluate the correctness of predicted actions by comparing it to the ground truth action with the action matching code released in the \aitw benchmark.
As suggested in the benchmark we compute the partial action matching score which is defined as the number of correct predictions divided by the episode length.
\citet{rawles2023androidinthewild} showed that the partial action matching score is correlated with the true complete match score which measures the ratio of tasks that were correctly fulfilled estimated by human evaluation.
As a result, the partial match score is an efficient and yet meaningful metric to measure the performance of an instruction-based UI agent.

\subsection{Comparative Evaluation}

We train our model on all training splits of the \aitw dataset jointly but similar to previous works \citep{zhan2023onlylookatscreens} use only $10\%$ of the data from the  \datasetgoogle split, as it is so much larger than the rest and would create a dataset imbalance.
To reduce hardware and computational requirements we train with LoRA. 
For the LoRA rank, we use a value of $32$, set $\alpha=64$, and use a dropout value of $0.05$. We use a batch size of $128$, a learning rate of $0.0003$, and trained for up to $5$ epochs.

We compare our method against strong baselines from several prior works: SeeClick \citep{cheng2024seeclick},
GPT4-V \citep{yan2023gpt4wonderland,gpt4vision},
MobileAgent \citep{ding2024mobileagent},
CogAgent \citep{hong2023cogagent},
AutoDroid \citep{wen2023empowering},
an LLM combined with a description of the screen from \citet{song2023navigating} which we term LLM+screen description,
and Auto-UI \citep{zhan2023onlylookatscreens}.
It is important to note that results obtained with Auto-UI cannot be directly compared due to its significant simplification of the action space for swipes, reducing it to four discrete possibilities: left, right, down, and up. This makes correct predictions according to the \aitw action checking substantially easier because of its focus on checking axial direction correctness. However, this representation falls short in real-world applications, lacking the ability to capture swipe magnitude or accommodate different directions such as diagonal swiping, which may be necessary, for instance, when dragging icons.
Opting for this simplified representation leads to overfitting on the \aitw evaluation protocol, and hence we refrain from using it despite some performance degradation on the benchmark.
We term our model using LLama and Vit \textit{LLama+ViT} and our VLM based on the Qwen-VL model \textit{UI-VLM}.

The findings from our evaluation, as summarized in Table \ref{table:partial_match}, offer several insights. 
Foremost, our most successful method achieves an overall accuracy of $78.9$\%, outperforming all other methods and establishing a new state-of-the-art for the \aitw benchmark. Notably, our approach even surpasses the performance of CogAgent, despite our VLM having considerably fewer parameters ($9.6$ billion compared to CogAgent's $18$ billion). We attribute this superiority to two pivotal factors.
First, by discretizing screen locations into categorical bins and using semantically meaningful words the language model can potentially better handle the actions encoded in textual form.
Additionally, unlike the other approaches, we incorporate a history of screenshots into our VLM. Previous research \citep{rawles2023androidinthewild} has demonstrated that a transformer model trained with both a history of screenshots and actions outperforms the same model trained solely on a single state input. This enhancement is logical, as certain scenarios may lack comprehensive information in a single screenshot, leading to partial observability issues. Using a history of observations mitigates this problem by providing context across multiple states.
For instance, in a webshop environment, where an item must be added to the cart before proceeding to checkout, a model without access to the history of screenshots may struggle to determine if the item has already been added.

Another noteworthy observation from our experiments is the superior performance of the Qwen-VL model compared to the LLama+ViT model. We hypothesize that pretraining the VLM on vision-language tasks before fine-tuning it as an agent contributes to this enhanced performance. Specifically, the Optical Character Recognition (OCR) pretraining of the Qwen-VL model likely positively impacts its downstream agent performance, particularly in the context of \aitw as for many tasks it is important to understand the text written on the screen.
Nevertheless, it is intriguing to note that the LLama+ViT model still achieves a respectable performance compared to other VLMs pretrained on vision-language tasks. This finding underscores the significant benefit of having a sequence of screenshots and highlights the effectiveness of our textual action representation, which the model easily comprehends.

\begin{table}[t!]
\centering
\scalebox{0.86}{
\begin{tabular}{l|l|lllll}
\toprule
Model & Overall & \datasetgoogle & \datasetinstall & \datasetshopping & \datasetgeneral & \datasetsingle \\
\hline
AutoDroid\footnote{According to the evaluation in \citet{song2023navigating}.}& 52.0 & - & - & - & - & - \\
LLM+screen description & 57.0 & - & - & - & - & - \\
GPT4-V    & 53.0 & 49.2 & 46.1 & 48.2 & 43.0 & 78.3 \\
SeeClick  & 59.8 & 57.7 & 64.5 & 57.3 & 56.0 & 63.6 \\
MobileAgent & 66.9 & 64.0 & 75.0 & 63.6 & 55.8 & 76.3  \\
Auto-UI\textbf{*}  & 74.3 & 71.4 & 76.9 & 70.3 & 68.2 & 84.6 \\
CogAgent & 76.9 & \textbf{74.9} & 78.9 & 71.7 & 65.3 & \textbf{93.5}  \\
\hline
LLama+VIT      &  73.7    &  70.8 & 77.1  & 68.2  & 69.3  & 83.1  \\
UI-VLM      &  \textbf{78.9}    & 72.4 & \textbf{83.3}  & \textbf{74.2}  & \textbf{74.7}  & 89.9  \\
\hline
\bottomrule
\end{tabular}
}
\caption{ Partial match scores across averaged over all splits and for each split separately. \textbf{*}Note that the results of Auto-UI are not directly comparable with our method because of a different action representation as explained in the main text. We still include those results for reference.}
\label{table:partial_match}
\end{table}
\section{Discussion}

In this study, we investigated the training of vision language models (VLMs) for controlling mobile devices based solely on their user interfaces. We evaluated different base VLMs and found that pretraining them on vision-language tasks before training them as agents yielded favorable results. Additionally, incorporating a history of observations and employing a simple yet semantically meaningful action space proved advantageous. Our most successful model achieves state-of-the-art performance on the \aitw benchmark.

Our findings shed light on effective design choices in training VLMs as agents. As language models continue to advance in capability, we anticipate that leveraging more powerful base VLMs will lead to further performance improvements when fine-tuned as agents. Tracking finetuning patterns that yield optimal agent performance will be crucial for maximizing the potential of VLMs, promising the development of highly useful agents.
As is it straightforward to transfer our approach to desktop computers,
it holds great promise to explore the possibilities within this domain in the future.

\newpage

\bibliography{iclr2024_conference}

\begin{thebibliography}{31}
\providecommand{\natexlab}[1]{#1}
\providecommand{\url}[1]{\texttt{#1}}
\expandafter\ifx\csname urlstyle\endcsname\relax
  \providecommand{\doi}[1]{doi: #1}\else
  \providecommand{\doi}{doi: \begingroup \urlstyle{rm}\Url}\fi

\bibitem[Achiam et~al.(2023)Achiam, Adler, Agarwal, Ahmad, Akkaya, Aleman, Almeida, Altenschmidt, Altman, Anadkat, et~al.]{achiam2023gpt}
Josh Achiam, Steven Adler, Sandhini Agarwal, Lama Ahmad, Ilge Akkaya, Florencia~Leoni Aleman, Diogo Almeida, Janko Altenschmidt, Sam Altman, Shyamal Anadkat, et~al.
\newblock Gpt-4 technical report.
\newblock \emph{arXiv preprint arXiv:2303.08774}, 2023.

\bibitem[Bai et~al.(2023{\natexlab{a}})Bai, Bai, Chu, Cui, Dang, Deng, Fan, Ge, Han, Huang, et~al.]{bai2023qwen}
Jinze Bai, Shuai Bai, Yunfei Chu, Zeyu Cui, Kai Dang, Xiaodong Deng, Yang Fan, Wenbin Ge, Yu~Han, Fei Huang, et~al.
\newblock Qwen technical report.
\newblock \emph{arXiv preprint arXiv:2309.16609}, 2023{\natexlab{a}}.

\bibitem[Bai et~al.(2023{\natexlab{b}})Bai, Bai, Yang, Wang, Tan, Wang, Lin, Zhou, and Zhou]{bai2023qwenvl}
Jinze Bai, Shuai Bai, Shusheng Yang, Shijie Wang, Sinan Tan, Peng Wang, Junyang Lin, Chang Zhou, and Jingren Zhou.
\newblock Qwen-vl: A frontier large vision-language model with versatile abilities.
\newblock \emph{arXiv preprint arXiv:2308.12966}, 2023{\natexlab{b}}.

\bibitem[Brohan et~al.(2023{\natexlab{a}})Brohan, Brown, Carbajal, Chebotar, Chen, Choromanski, Ding, Driess, Dubey, Finn, et~al.]{brohan2023rt2}
Anthony Brohan, Noah Brown, Justice Carbajal, Yevgen Chebotar, Xi~Chen, Krzysztof Choromanski, Tianli Ding, Danny Driess, Avinava Dubey, Chelsea Finn, et~al.
\newblock Rt-2: Vision-language-action models transfer web knowledge to robotic control.
\newblock \emph{arXiv preprint arXiv:2307.15818}, 2023{\natexlab{a}}.

\bibitem[Brohan et~al.(2023{\natexlab{b}})Brohan, Chebotar, Finn, Hausman, Herzog, Ho, Ibarz, Irpan, Jang, Julian, et~al.]{saycan}
Anthony Brohan, Yevgen Chebotar, Chelsea Finn, Karol Hausman, Alexander Herzog, Daniel Ho, Julian Ibarz, Alex Irpan, Eric Jang, Ryan Julian, et~al.
\newblock Do as i can, not as i say: Grounding language in robotic affordances.
\newblock In \emph{Conference on Robot Learning}, pp.\  287--318. PMLR, 2023{\natexlab{b}}.

\bibitem[Cheng et~al.(2024)Cheng, Sun, Chu, Xu, Li, Zhang, and Wu]{cheng2024seeclick}
Kanzhi Cheng, Qiushi Sun, Yougang Chu, Fangzhi Xu, Yantao Li, Jianbing Zhang, and Zhiyong Wu.
\newblock Seeclick: Harnessing gui grounding for advanced visual gui agents.
\newblock \emph{arXiv preprint arXiv:2401.10935}, 2024.

\bibitem[Ding(2024)]{ding2024mobileagent}
Tinghe Ding.
\newblock Mobileagent: enhancing mobile control via human-machine interaction and sop integration.
\newblock \emph{arXiv preprint arXiv:2401.04124}, 2024.

\bibitem[Dosovitskiy et~al.(2020)Dosovitskiy, Beyer, Kolesnikov, Weissenborn, Zhai, Unterthiner, Dehghani, Minderer, Heigold, Gelly, et~al.]{dosovitskiy2020image}
Alexey Dosovitskiy, Lucas Beyer, Alexander Kolesnikov, Dirk Weissenborn, Xiaohua Zhai, Thomas Unterthiner, Mostafa Dehghani, Matthias Minderer, Georg Heigold, Sylvain Gelly, et~al.
\newblock An image is worth 16x16 words: Transformers for image recognition at scale.
\newblock \emph{arXiv preprint arXiv:2010.11929}, 2020.

\bibitem[Driess et~al.(2023)Driess, Xia, Sajjadi, Lynch, Chowdhery, Ichter, Wahid, Tompson, Vuong, Yu, et~al.]{driess2023palm}
Danny Driess, Fei Xia, Mehdi~SM Sajjadi, Corey Lynch, Aakanksha Chowdhery, Brian Ichter, Ayzaan Wahid, Jonathan Tompson, Quan Vuong, Tianhe Yu, et~al.
\newblock Palm-e: An embodied multimodal language model.
\newblock \emph{arXiv preprint arXiv:2303.03378}, 2023.

\bibitem[GeminiTeam et~al.(2023)GeminiTeam, Anil, Borgeaud, Wu, Alayrac, Yu, Soricut, Schalkwyk, Dai, Hauth, et~al.]{team2023gemini}
Google GeminiTeam, Rohan Anil, Sebastian Borgeaud, Yonghui Wu, Jean-Baptiste Alayrac, Jiahui Yu, Radu Soricut, Johan Schalkwyk, Andrew~M Dai, Anja Hauth, et~al.
\newblock Gemini: a family of highly capable multimodal models.
\newblock \emph{arXiv preprint arXiv:2312.11805}, 2023.

\bibitem[Hong et~al.(2023)Hong, Wang, Lv, Xu, Yu, Ji, Wang, Wang, Dong, Ding, et~al.]{hong2023cogagent}
Wenyi Hong, Weihan Wang, Qingsong Lv, Jiazheng Xu, Wenmeng Yu, Junhui Ji, Yan Wang, Zihan Wang, Yuxiao Dong, Ming Ding, et~al.
\newblock Cogagent: A visual language model for gui agents.
\newblock \emph{arXiv preprint arXiv:2312.08914}, 2023.

\bibitem[Hu et~al.(2021)Hu, Wallis, Allen-Zhu, Li, Wang, Wang, Chen, et~al.]{hu2021lora}
Edward~J Hu, Phillip Wallis, Zeyuan Allen-Zhu, Yuanzhi Li, Shean Wang, Lu~Wang, Weizhu Chen, et~al.
\newblock Lora: Low-rank adaptation of large language models.
\newblock In \emph{International Conference on Learning Representations}, 2021.

\bibitem[Ilharco et~al.(2021)Ilharco, Wortsman, Wightman, Gordon, Carlini, Taori, Dave, Shankar, Namkoong, Miller, Hajishirzi, Farhadi, and Schmidt]{openclip}
Gabriel Ilharco, Mitchell Wortsman, Ross Wightman, Cade Gordon, Nicholas Carlini, Rohan Taori, Achal Dave, Vaishaal Shankar, Hongseok Namkoong, John Miller, Hannaneh Hajishirzi, Ali Farhadi, and Ludwig Schmidt.
\newblock Openclip, July 2021.
\newblock URL \url{https://doi.org/10.5281/zenodo.5143773}.
\newblock If you use this software, please cite it as below.

\bibitem[Li et~al.(2023)Li, Li, Savarese, and Hoi]{li2023blip2}
Junnan Li, Dongxu Li, Silvio Savarese, and Steven Hoi.
\newblock Blip-2: Bootstrapping language-image pre-training with frozen image encoders and large language models.
\newblock \emph{arXiv preprint arXiv:2301.12597}, 2023.

\bibitem[Lifshitz et~al.(2023)Lifshitz, Paster, Chan, Ba, and McIlraith]{lifshitz2023steve}
Shalev Lifshitz, Keiran Paster, Harris Chan, Jimmy Ba, and Sheila McIlraith.
\newblock Steve-1: A generative model for text-to-behavior in minecraft.
\newblock \emph{arXiv preprint arXiv:2306.00937}, 2023.

\bibitem[Nakajima(2023)]{babyagi}
Yohei Nakajima.
\newblock Babyagi, 2023.
\newblock URL \url{https://github.com/yoheinakajima/babyagi}.

\bibitem[OpenAI(2023)]{gpt4vision}
OpenAI.
\newblock Gpt-4v(ision) system card.
\newblock 2023.
\newblock URL \url{https://cdn.openai.com/papers/GPTV_System_Card.pdf}.

\bibitem[Parisi et~al.(2022)Parisi, Zhao, and Fiedel]{parisi2022talm}
Aaron Parisi, Yao Zhao, and Noah Fiedel.
\newblock Talm: Tool augmented language models.
\newblock \emph{arXiv preprint arXiv:2205.12255}, 2022.

\bibitem[Patil et~al.(2023)Patil, Zhang, Wang, and Gonzalez]{patil2023gorilla}
Shishir~G Patil, Tianjun Zhang, Xin Wang, and Joseph~E Gonzalez.
\newblock Gorilla: Large language model connected with massive apis.
\newblock \emph{arXiv preprint arXiv:2305.15334}, 2023.

\bibitem[Qin et~al.(2023)Qin, Liang, Ye, Zhu, Yan, Lu, Lin, Cong, Tang, Qian, et~al.]{qin2023toolllm}
Yujia Qin, Shihao Liang, Yining Ye, Kunlun Zhu, Lan Yan, Yaxi Lu, Yankai Lin, Xin Cong, Xiangru Tang, Bill Qian, et~al.
\newblock Toolllm: Facilitating large language models to master 16000+ real-world apis.
\newblock \emph{arXiv preprint arXiv:2307.16789}, 2023.

\bibitem[Rawles et~al.(2023)Rawles, Li, Rodriguez, Riva, and Lillicrap]{rawles2023androidinthewild}
Christopher Rawles, Alice Li, Daniel Rodriguez, Oriana Riva, and Timothy~P Lillicrap.
\newblock Androidinthewild: A large-scale dataset for android device control.
\newblock In \emph{Thirty-seventh Conference on Neural Information Processing Systems Datasets and Benchmarks Track}, 2023.

\bibitem[Reworkd(2023)]{agentgpt}
Reworkd.
\newblock Agentgpt, 2023.
\newblock URL \url{https://github.com/reworkd/AgentGPT}.

\bibitem[Richards(2023)]{autogpt}
Toran~Bruce Richards.
\newblock Auto-gpt: An autonomous gpt-4 experiment, 2023.
\newblock URL \url{https://github.com/Significant-Gravitas/Auto-GPT}.

\bibitem[Schick et~al.(2023)Schick, Dwivedi-Yu, Dess{\`\i}, Raileanu, Lomeli, Zettlemoyer, Cancedda, and Scialom]{schick2023toolformer}
Timo Schick, Jane Dwivedi-Yu, Roberto Dess{\`\i}, Roberta Raileanu, Maria Lomeli, Luke Zettlemoyer, Nicola Cancedda, and Thomas Scialom.
\newblock Toolformer: Language models can teach themselves to use tools.
\newblock \emph{arXiv preprint arXiv:2302.04761}, 2023.

\bibitem[Song et~al.(2023)Song, Bian, Tang, and Cai]{song2023navigating}
Yunpeng Song, Yiheng Bian, Yongtao Tang, and Zhongmin Cai.
\newblock Navigating interfaces with ai for enhanced user interaction.
\newblock \emph{arXiv preprint arXiv:2312.11190}, 2023.

\bibitem[Touvron et~al.(2023)Touvron, Martin, Stone, Albert, Almahairi, Babaei, Bashlykov, Batra, Bhargava, Bhosale, et~al.]{touvron2023llama2}
Hugo Touvron, Louis Martin, Kevin Stone, Peter Albert, Amjad Almahairi, Yasmine Babaei, Nikolay Bashlykov, Soumya Batra, Prajjwal Bhargava, Shruti Bhosale, et~al.
\newblock Llama 2: Open foundation and fine-tuned chat models.
\newblock \emph{arXiv preprint arXiv:2307.09288}, 2023.

\bibitem[Wang et~al.(2023{\natexlab{a}})Wang, Xie, Jiang, Mandlekar, Xiao, Zhu, Fan, and Anandkumar]{wang2023voyager}
Guanzhi Wang, Yuqi Xie, Yunfan Jiang, Ajay Mandlekar, Chaowei Xiao, Yuke Zhu, Linxi Fan, and Anima Anandkumar.
\newblock Voyager: An open-ended embodied agent with large language models.
\newblock \emph{arXiv preprint arXiv:2305.16291}, 2023{\natexlab{a}}.

\bibitem[Wang et~al.(2023{\natexlab{b}})Wang, Lv, Yu, Hong, Qi, Wang, Ji, Yang, Zhao, Song, et~al.]{wang2023cogvlm}
Weihan Wang, Qingsong Lv, Wenmeng Yu, Wenyi Hong, Ji~Qi, Yan Wang, Junhui Ji, Zhuoyi Yang, Lei Zhao, Xixuan Song, et~al.
\newblock Cogvlm: Visual expert for pretrained language models.
\newblock \emph{arXiv preprint arXiv:2311.03079}, 2023{\natexlab{b}}.

\bibitem[Wen et~al.(2023)Wen, Li, Liu, Zhao, Yu, Li, Jiang, Liu, Zhang, and Liu]{wen2023empowering}
Hao Wen, Yuanchun Li, Guohong Liu, Shanhui Zhao, Tao Yu, Toby Jia-Jun Li, Shiqi Jiang, Yunhao Liu, Yaqin Zhang, and Yunxin Liu.
\newblock Empowering llm to use smartphone for intelligent task automation.
\newblock \emph{arXiv preprint arXiv:2308.15272}, 2023.

\bibitem[Yan et~al.(2023)Yan, Yang, Zhu, Lin, Li, Wang, Yang, Zhong, McAuley, Gao, et~al.]{yan2023gpt4wonderland}
An~Yan, Zhengyuan Yang, Wanrong Zhu, Kevin Lin, Linjie Li, Jianfeng Wang, Jianwei Yang, Yiwu Zhong, Julian McAuley, Jianfeng Gao, et~al.
\newblock Gpt-4v in wonderland: Large multimodal models for zero-shot smartphone gui navigation.
\newblock \emph{arXiv preprint arXiv:2311.07562}, 2023.

\bibitem[Zhan \& Zhang(2023)Zhan and Zhang]{zhan2023onlylookatscreens}
Zhuosheng Zhan and Aston Zhang.
\newblock You only look at screens: Multimodal chain-of-action agents.
\newblock \emph{arXiv preprint arXiv:2309.11436}, 2023.

\end{thebibliography}
\bibliographystyle{iclr2024_conference}

\newpage
\appendix

\section{Android in the Wild Benchmark}

\vspace{-0.2cm}
\aitw stands as a comprehensive benchmark dataset designed for UI control, encompassing natural language instructions, screenshots, and actions. The dataset comprises 715K episodes across 30K distinct instructions, covering a wide array of multi-step tasks such as navigating applications, web searches, and online shopping, spanning over 350 apps and websites. The demonstrations are recorded on various operating systems and device types with differing screen resolutions.

The benchmark dataset comprises five subsets: \datasetgoogle, \datasetinstall, \datasetshopping, \datasetgeneral, and \datasetsingle, each serving different purposes.  Each subset is divided into training, validation, and test sets based on episodes, with proportions of 80\%/10\%/10\%.

To evaluate if a predicted action is correct we compare it to the ground truth action with the action matching code released in the \aitw benchmark.
Actions are deemed to match if their types are identical. When it comes to dual-point taps, we equate them if they occur within 14\% of the screen distance from one another. Alternatively, if tap actions happen within the confines of the same detected bounding box (expanded to 240\% of its original size during action matching), they are regarded as equal. Lastly, two dual-point scrolls are considered equal if they share the same primary scroll axis, be it vertical or horizontal.

As suggested in the \aitw benchmark we compute the partial action matching score which is defined as the number of correct predictions divided by the episode length.
\citet{rawles2023androidinthewild} showed that the partial action matching score is correlated with the true complete match score which measures the ratio of tasks that were correctly fulfilled estimated by human evaluation.
As a result, the partial match score is an efficient and yet meaningful metric to measure the performance of an instruction-based UI agent.

\section{Detailed Action Representation}
In Table \ref{table:partial_match} we give the detailed mapping between the actions in \aitw and its representation in natural language we use for the VLM. The \emph{BIN} denotes the map of the normalized actions interval $[0,1]$ to equally spaced bins represented by integers from $0$ to $99$. 
\begin{table}[h]
\centering
\scalebox{0.74}{
\begin{tabular}{l|l|}
\toprule
\aitw action & natural language representation  \\
\hline
\\
\{'action\_type': DUAL\_POINT, 'touch\_yx': [ty, tx], 'lift\_yx': None, 'action\_text': None\}  & tap at \emph{BIN}(ty) \emph{BIN}(tx)\\
\multirow{2}{*}{
\{'action\_type': DUAL\_POINT, 'touch\_yx': [ty, tx], 'lift\_yx': [ly, lx], 'action\_text': None\} 
} & swipe from \emph{BIN}(ty) \emph{BIN}(tx) to \\ & \emph{BIN}(ly) \emph{BIN}(lx)\\
\{'action\_type': TYPE, 'touch\_yx': None, 'lift\_yx': None, 'action\_text': \emph{some text}\}  & Input text "\emph{some text}" \\
\{'action\_type': PRESS\_BACK, 'touch\_yx': None, 'lift\_yx': None, 'action\_text': None\}  & press back \\
\{'action\_type': PRESS\_HOME, 'touch\_yx': None, 'lift\_yx': None, 'action\_text': None\}  & press home \\
\{'action\_type': PRESS\_ENTER, 'touch\_yx': None, 'lift\_yx': None, 'action\_text': None\}  & press enter \\
\{'action\_type': TASK\_COMPLETE, 'touch\_yx': None, 'lift\_yx': None, 'action\_text': None\}  & complete \\
\{'action\_type': TASK\_IMPOSSIBLE, 'touch\_yx': None, 'lift\_yx': None, 'action\_text': None\}  & impossible \\
 \\
\bottomrule
\end{tabular}
}
\caption{Mapping between an action in \aitw and its represenation in natural language.}
\label{table:action_map}
\end{table}

\end{document}